\pdfoutput=1

\documentclass[11pt]{article}

\usepackage[]{naacl2021}

\usepackage{times}
\usepackage{latexsym}

\usepackage[T1]{fontenc}

\usepackage[utf8]{inputenc}

\usepackage{microtype}

\usepackage{multirow}  
\usepackage{amssymb} 
\usepackage{graphicx}  
\usepackage{amsmath}
\usepackage{subfigure}
\usepackage{makecell}
\usepackage{booktabs}
\usepackage{enumitem}

\title{Memory Augmented Sequential Paragraph Retrieval for Multi-hop Question Answering}

\author{
	Nan Shao$^\dag$,
  	Yiming Cui$^\ddag$$^\dag$,
  	\textbf{Ting Liu$^\ddag$},
  	\textbf{Shijin Wang$^\dag$$^\S$},
  	\textbf{Guoping Hu$^\dag$}\\
  	{$^\dag$State Key Laboratory of Cognitive Intelligence, iFLYTEK Research, China} \\
  	{$^\ddag$Research Center for Social Computing and Information Retrieval (SCIR), } \\
  	{Harbin Institute of Technology, Harbin, China} \\
  	{$^\S$iFLYTEK AI Research (Hebei), Langfang, China} \\
  	$^\dag$$^\S$\tt\{nanshao,ymcui,sjwang3,gphu\}@iflytek.com \\
  	$^\ddag$\tt\{ymcui,tliu\}@ir.hit.edu.cn
  }

\begin{document}
\maketitle
\begin{abstract}
Retrieving information from correlative paragraphs or documents to answer open-domain multi-hop questions is very challenging.
To deal with this challenge, most of the existing works consider paragraphs as nodes in a graph and propose graph-based methods to retrieve them. 
However, in this paper, we point out the intrinsic defect of such methods.
Instead, we propose a new architecture that models paragraphs as sequential data and considers multi-hop information retrieval as a kind of sequence labeling task.
Specifically, we design a rewritable external memory to model the dependency among paragraphs. Moreover, a threshold gate mechanism is proposed to eliminate the distraction of noise paragraphs.
We evaluate our method on both full wiki and distractor subtask of HotpotQA, a public textual multi-hop QA dataset requiring multi-hop information retrieval. Experiments show that our method achieves significant improvement over the published state-of-the-art method in retrieval and downstream QA task performance.

\end{abstract}

\section{Introduction}
\label{sec:introduction}
Open-domain Question Answering (QA) is a popular topic in natural language processing that requires models to answer questions given a large collection of text paragraphs (e.g., Wikipedia). Most previous works leverage a retrieval model to calculate semantic similarity between a question and each paragraph to retrieve a set of paragraphs, then a reading comprehension model extracts an answer from one of the paragraphs.
These pipeline methods work well in single-hop question answering, where the answer can be derived from only one paragraph. However, many real-world questions produced by users are multi-hop questions that require reasoning across multiple documents or paragraphs. 

\begin{figure}
	\centering
	\includegraphics[width=7.5cm]{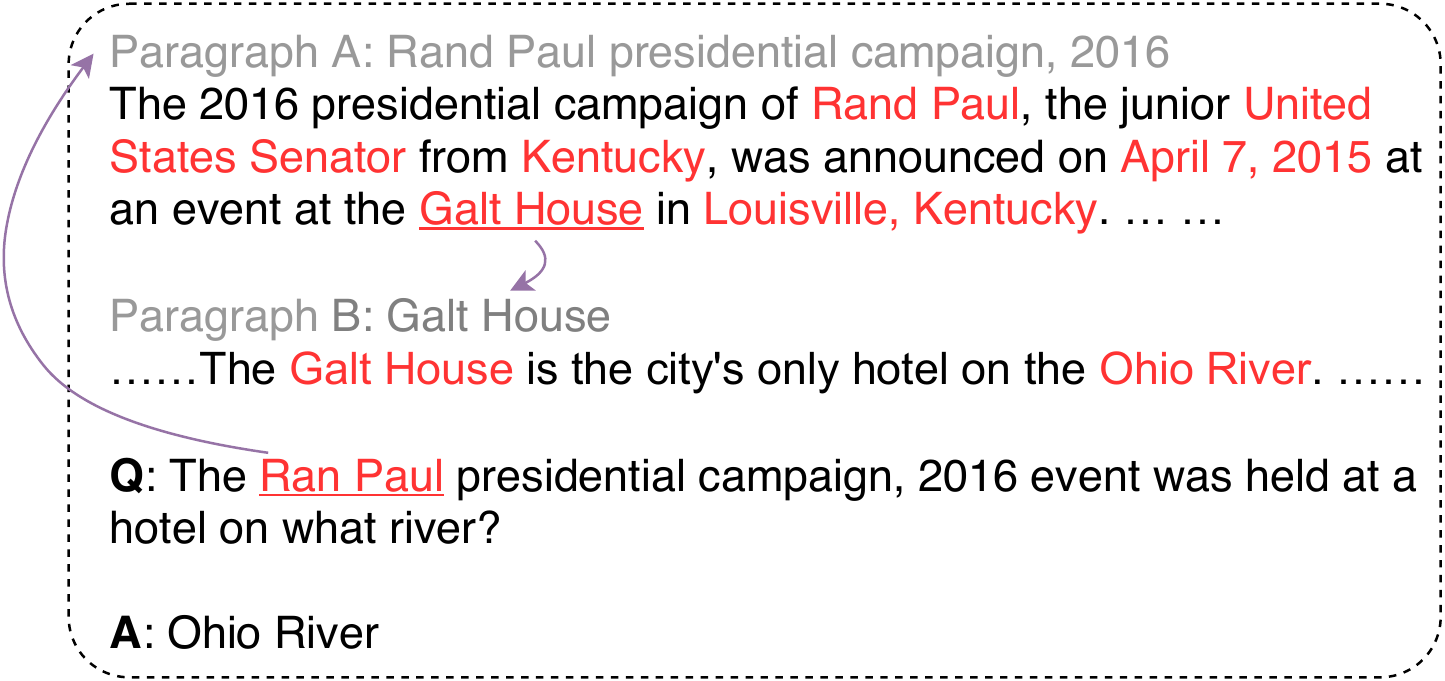}
	\caption{\label{fig: example}
	An example of open-domain multi-hop question from HotpotQA. The model needs to retrieve evidence paragraphs from entire Wikipedia and derive the answer.
	}
\end{figure}

In this paper, we study the problem of textual multi-hop question answering at scale, which requires multi-hop information retrieval. 
An example from HotpotQA is illustrated in Figure \ref{fig: example}. To answer the question \textit{``The Rand Paul presidential campaign, 2016 event was held at a hotel on what river?''}, the retrieval model needs to first identify \textit{Paragraph A} as an evidence paragraph according to the similarity of the terms. 
The retrieval model learns that the event was held at a hotel named \textit{``Galt House''}, which leads to the next-hop \textit{Paragraph B}. 
Existing neural-based or non-neural based methods can not perform well for such questions due to little lexical overlap or semantic relations between the question and \textit{Paragraph B}.

To tackle this challenge, the mainstream studies model all paragraphs as a graph, where paragraphs are connected if they have the same entity mentions or hyperlink relation \cite{ding-etal-2019-cognitive,Zhao2020Transformer-XH,Asai2020Learning}. For example, Graph-based Recurrent Retriever \cite{Asai2020Learning} first leverage non-parameterized methods (e.g., TF-IDF or BM25) to retrieve a set of initial paragraphs as starting points, then a neural retrieval model will determine whether paragraphs that linked to an initial paragraph is the next step of reasoning path. However, the method suffers several intrinsic drawbacks:
(1) These graph-based methods are based on a hypothesis that evident paragraphs should have the same entity mentions or linked by a hyperlink so that they can perform well on bridge questions. However, for comparison questions (e.g., does A and B have the same attribute?), the evidence paragraphs about A and B are independent. 
(2) Once a paragraph is identified as an evidence paragraph, graph-based retriever like the Graph-based Recurrent Retriever \cite{Asai2020Learning} will continue to determine whether the paragraphs linked from the evidence paragraph are relevant to the question, which implies that these models have to assume the relation between paragraphs is directional.
(3) Several graph-based methods have to process all adjacent paragraphs simultaneously, leading to inefficient memory usage.

In this paper, we discard the mainstream practice. Instead, we propose to treat all candidate paragraphs as sequential data and identify them iteratively. The newly proposed framework for solving multi-hop information retrieval has several desirable advantages: 
(1) The framework does not assume entity mentions or hyperlinks between two evidence paragraphs so that it can be suitable for both bridge and comparison questions.
(2) It is natural for the framework to take all paragraphs connected to or from a certain paragraph into consideration.
(3) As only one paragraph is located in GPU memory at the same time, the framework is memory efficient.

We implement the framework and propose the Gated Memory Flow model. Inspired by Neural Turning Machine \cite{graves2014neural}, we leverage an external rewritable memory to memorizes information extract from previous paragraphs. The model iteratively reads a paragraph and the memory to identify whether the current paragraph is an evidence paragraph. Due to many noise paragraphs in the candidate set, we design a threshold gating mechanism to control the writing procedure. In practice, we find that different negative examples in the training set affect the retrieval performance significantly and propose a simple method to find more valuable negatives for the retrieval model. 

Our method significantly outperforms previous methods on HotpotQA under both the full wiki and the distractor settings. In analysis experiments, we demonstrate the effectiveness of our method and how different settings in the model affect downstream performance.

Our contributions are summarized as follows:
\begin{itemize}
	\item We propose a new framework that treats paragraphs as sequential data and iteratively retrieves them for solving multi-hop questing answering at scale.
	\item We implement the framework and propose the Gated Memory Flow model. Besides, we introduce a simple method to find more valuable negatives for open-domain multi-hop QA. 
	\item Our method significantly outperforms all the published methods on HotpotQA 
by a large margin. Extensive experiments demonstrate the effectiveness of our method.
\end{itemize}

\begin{figure*}
	\centering
	\includegraphics[width = 15cm]{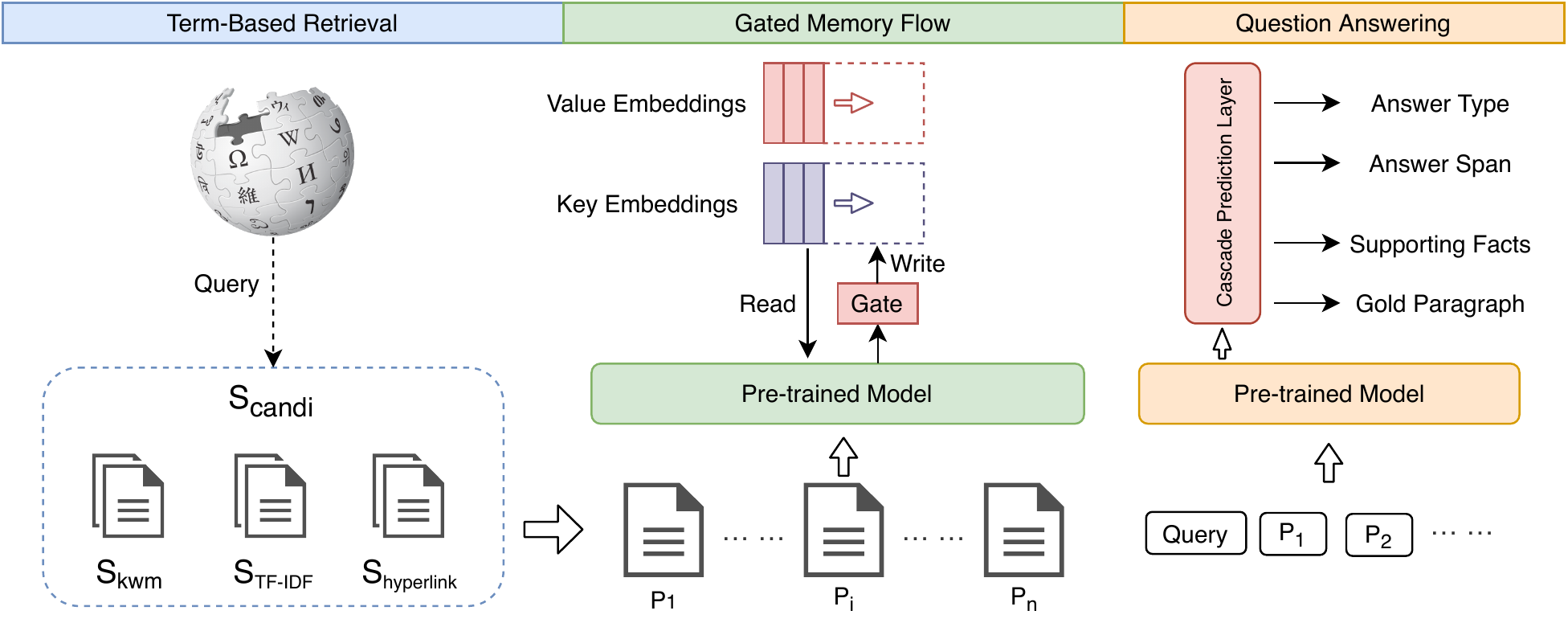}
	\caption{\label{fig: overview}
	Overview of our architecture.}
\end{figure*}

\section{Related Work}
\label{sec:related}
\paragraph{Textual Multi-hop Question Answering.}
In contrast to single-hop question answering, multi-hop question answering requires model reasoning across multiple documents or paragraphs to derive the answer. 
In recent years, this topic has attracted many researchers attention, and many datasets have been released \cite{welbl-etal-2018-constructing, talmor-berant-2018-web, yang-etal-2018-hotpotqa, Khot2020QASCAD, xie-etal-2020-worldtree}.
In this paper, we focus our study on textual on the textual multi-hop QA dataset, HotpotQA.

Methods for solving multi-hop QA can be roughly divided into two categories: graph-based methods and query reformulation based methods. 
Graph-based methods often construct an entity graph based on co-reference resolution or co-occurrence \cite{dhingra-etal-2018-neural,song2018exploring,de-cao-etal-2019-question}. These works show the GNN frameworks like graph convolution network \cite{kipf2017semi} and graph attention network \cite{velivckovic2018graph} perform on an entity graph can achieve promising results in WikiHop. Tu~\shortcite{tu-etal-2019-multi} propose to model paragraphs and candidate answers as different kinds of nodes in a graph and extend an entity graph to a heterogeneous graph. 
In order to adapt such methods to span-based QA tasks like HotpotQA, DFGN \cite{qiu-etal-2019-dynamically} leverage a dynamical fusion layer to combine graph representations and sequential text representations.

Query reformulation based methods solve multi-hop QA tasks by implicitly or explicitly reformulating the query at different reasoning hop. For example, QFE \cite{nishida-etal-2019-answering} update query representations at different hop. DecompRC \cite{min-etal-2019-multi} decomposes a compositional question into several single-hop questions.

\paragraph{Multi-hop Information Retrieval.} Chen~\shortcite{chen-etal-2017-reading} first propose to leverage the entire Wikipedia paragraphs to answer open-domain questions. Most existing open-domain QA methods use a pipeline approach that includes a retriever model and a reader model. For open-domain multi-hop QA, graph-based methods are also mainstream practice. These studies organized paragraphs as nodes in a graph structure and connected them according to hyperlinks.  Cognitive Graph \cite{ding-etal-2019-cognitive} employ a reading comprehension model to predict answer spans and next-hop answer to construct a cognitive graph. Transformer-XH \cite{Zhao2020Transformer-XH} introduce eXtra Hop attention to reasons over a paragraph graph and produces the global representation of all paragraphs to predict the answers. Due to encoding paragraphs independently, this method lacks fine-grained evidence paragraphs interaction and memory-inefficient. Asai~\shortcite{Asai2020Learning} proposes a graph-based recurrent model to retrieves a reasoning chain through a path in the paragraph graph. As described in Section \ref{sec:introduction}, this method may not perform well on comparison questions and suffer from directional edges caused low recall.

 There are also several studies propose query reformulation based methods for multi-hop information retrieval. Multi-step Reasoner \cite{das2018multistep} employees a retriever interacts with a reader model and reformulates the query in a latent space for the next retrieval hop. GoldEn Retriever \cite{qi-etal-2019-answering} generates several single-hop questions at different hop. 
 These methods often perform poorly in result for the error accumulation through different hops.

\section{Method}
In this section, we introduce the pipeline we designed for multi-hop QA, the overview of our system is shown in Figure \ref{fig: overview}. 
We first leverage a heuristic term-based retrieval method to find candidate paragraphs for the question $Q$ as much as possible. Our Gated Memory Flow model takes all candidate paragraphs as input and processes them one by one. Finally, paragraphs that gain the highest relevance score for a question are selected and fed into a neural question answering model. The reader model uses a cascade prediction layer to predict all targets in a multi-task way.

\subsection{Term-Based Retrieval}
We leverage a heuristic term-based retrieval method to construct the candidate paragraphs set that contains all possible paragraphs. Paragraphs in the candidate set come from three sources.
\paragraph{Key Word Matching.} We retrieve all paragraphs whose title exact match terms in the query. The top $N_{kwm}$ paragraphs with highest TF-IDF score are collected to set $P_{kwm}$.
\paragraph{TF-IDF.} The top $N_{tfidf}$ paragraphs retrieved by DrQA's TF-IDF on the question \cite{chen-etal-2017-reading}. 
\paragraph{Hyperlink.} Paragraphs that necessary for derived the answer but do not have lexical overlap to the question often have hyperlink relation to the paragraphs retrieved by key work matching or TF-IDF. Therefore, for each paragraph $P_i$ in set $P_{kwm}$ and $P_{tfidf}$, we extract all the hyperlinked paragraphs from $P_i$ to construct set $P_{hyperlink}$. Unlike previous works \cite{nie-etal-2019-revealing, Asai2020Learning}, both paragraphs that have a hyperlink connect to $P_i$ and paragraphs connected from $P_i$ are taken into consideration.

Finally, the three collections are merged into a candidate set $P_{cand}$.

\subsection{Gated Memory Flow}
After retrieving the candidate set, we propose a neural-based model named Gated Memory Flow that accurately selects evidence paragraphs from the candidate set. 
\paragraph{Model.} As describe above, for a compositional question, whether a paragraph contains evidence information is not only dependent on the query but also other paragraphs. The task can be formulated as:
\begin{equation}
	\mathop{\arg\max}_{\theta}P(P_t\in\mathbf{E}|Q, P_{1:t-1}, \theta)
\end{equation}
where the $\mathbf{E}$ denotes evidence set, $\theta$ denotes the parameters of the model, $P_t$ is the $t$-th paragraphs in candidate set $P_{cand}=\{P_1, \dots, P_t, \dots, P_n\}$ and $P_{1:t-1}$ represents the processed $t-1$ paragraphs.

At the $t$-th time step, the model takes paragraph $P_t$ as input and determines whether it is an evidence paragraph depending on identified paragraphs. The question $Q$ and paragraph $P_t$ are concatenated and fed into a pre-trained model to get the query-aware paragraph representations $\mathbf{H}_{t} \in \mathbb{R}^{l \times d}$.
To decrease parameters, we compress the matrix into a vector representation, then we employ the vector as a query vector to address desired information from the external memory module.
\begin{equation}
	\mathbf{x}_{t} = MeanPooling(\mathbf{H}_{t}) \in \mathbb{R}^{d}
\end{equation}

We denote the external memory at step $t$ as $\mathbf{M}_t \in \mathbb{R}^{l_m \times d}$, where $l_m$ is number of memory slots that have stored information at step $t$. We denote the readout vector representation at step $t$ as $\mathbf{o}_t$, which include the missing information to identify the current paragraph. The $\mathbf{o}_t$ and $\mathbf{x}_{t}$ are used to identify  
$D_t$:
\begin{align}
	\mathbf{h}_{t} &= tanh(\mathbf{W}_o\cdot \mathbf{o}_{t}+ \mathbf{W}\cdot \mathbf{x}_t) \in \mathbb{R}^{d} \\
	s_t &= \sigma(\mathbf{W}_s\cdot \mathbf{h}_{t})
\end{align}
where $s_t$ represent the relevant score between paragraph $P_t$ and question $Q$.

Now we describe the memory module and read/write procedure in detail. 
We follow the KVMemNN \cite{miller-etal-2016-key} architecture, which defines the memory slots as pairs of vectors $\{(\mathbf{k}_1, \mathbf{v}_1), \dots , (\mathbf{k}_{l_m}, \mathbf{v}_{l_m})\}$. In our implementation, the key vectors and value vectors are the same. Only a part of vectors are necessary for identifying $P_t$, therefore we use key vectors and query vector to address the memory. The addressing and reading procedure are similar to soft attention:
\begin{align}
	p_i &= Softmax(\mathbf{W}_q \mathbf{x}_t \cdot \mathbf{W}_k\mathbf{k}_i) \\
	\mathbf{o}_t &= \sum_{i}p_i\mathbf{W}_v\mathbf{v}_i
\end{align}
where $p_i$ denotes the readout probability of $i$-th memory slot. In our implementation, we extend the attention readout mechanism to a multi-head way \cite{vaswani2017attention}:
\begin{equation}
	\mathbf{o}_t = \textmd{Concat}(\mathbf{o}^{(1)}_t, \dots, \mathbf{o}^{(h)}_t)
\end{equation}
$\mathbf{o}^{(h)}_t$ denotes the output of $h$-th readout head. Due to many irrelevant paragraphs in the candidate set, we leverage a threshold gating mechanism to control the writing permission. The writing operation is concatenation for keeping the model concise. 
\begin{equation}
\mathbf{M}_{t+1}=
\begin{cases}
\mathbf{M}_{t} & s_t < gate\\
\textmd{Concat}(\mathbf{M}_{t}, x_t) & s_t \geq gate
\end{cases}
\end{equation}
The value of scalar $gate$ is a pre-defined hyper-parameter.

\paragraph{Training.} The parameters in GMF were updated with a binary cross entropy loss function:
\begin{equation}
\begin{aligned}
	\mathcal{L}_{retri}= -&\sum_{P_t\in P_{pos}}log(s_t) \\
	-&\sum_{P_t\in P_{neg}}log(1-s_t)
\end{aligned}
\end{equation}
where $P_{neg}$ contains eight negative paragraphs sampled by our negative sampling strategy for each question. At the beginning of training, the model can not give each paragraph an accurate relevant score, leading to unexpected behavior of the gate. For this reason, we only activate the gate after the first epoch of training.

\paragraph{Training with Harder Negative Examples.}
For each question in the training stage, we pass two gold paragraphs and eight negative examples from the term-based retrieval result to Gated Memory Flow. Intuitively, different negative sampling strategies may influence the training result, but previous works do not explore the effect.

We empirically demonstrate harder negative examples lead to better training results and propose a simple but effective strategy to find them (see Sec. \ref{sec:analysis}). Specifically, we first train a BERT-based binary classifier to calculate the relevant score between the questions and paragraphs. Eight paragraphs are randomly selected from term based retrieval results, combined with two evidence paragraphs for each question. Then we employ the model as a ranker to select top-8 paragraphs as augmented negative examples. We train our GMF model with these augmented examples. 

\paragraph{Inference.} At the inference stage, there may be hundreds of candidate paragraphs for a question. Nevertheless, it is unnecessary to put all of them in GPU memory besides the parameters of the model and the current paragraph $P_t$. To balance inference speed and memory usage, we load a \textit{chunk} of paragraphs into GPU memory at once.
After the model gives all paragraphs a relevant score, we sort them by the score and retrieve the top $N_{retri}$ paragraphs with the scores higher than a threshold $h_d$ for each question. We take these paragraphs as the input for the reader model.

\subsection{Reader Model}
There are many existing readers model for multi-hop QA, most of which use a graph-based approach. However, a recent study \cite{shao2020graph} shows self-attention in the pre-trained models can perform well in multi-hop QA. Therefore, we employ a pre-trained model as a reader to solve multi-hop reasoning.

We combine all the paragraphs selected by Gated Memory Flow into context $C$. For each example, we concatenate the question $Q$ and context $C$ as input and fed into a pre-trained model following a projection layer.

We follow the similar structure of the prediction layer from Yang~\shortcite{yang-etal-2018-hotpotqa}. There are four sub-tasks for the prediction layer: (1) evidence paragraphs prediction as an auxiliary task; (2) supporting facts prediction based on evidence paragraphs; (3) the start and end positions of span answer; (4) answer type prediction. We use a cascade structure due to the dependency between different outputs. Five isomorphic BiLSTMs are stacked layer by layer to obtain different representations for each sub-task. To predict evidence paragraphs, we take the first and last token of $i$-th paragraph as its representations.
\begin{align}
	\mathbf{M}_{p} &= {\rm \textmd{BiLSTM}}(\mathbf{C}) \in \mathbb{R}^{m \times d_2} \\
	\mathbf{O}_{para}&=\sigma(\mathbf{W}_p[\mathbf{M}_{p}[P^{(i)}_{s}]; \mathbf{M}_{p}[P^{(i)}_{e}]])
\end{align}
where $P^{(i)}_{s}$ and $P^{(i)}_{e}$ denote the start and end token index of $i$-th paragraph. Then we can predict whether the $i$-th sentence is a supporting fact depend on the outputs of evidence paragraph prediction.
\begin{align}
	\mathbf{M}_{s} &= {\rm \textmd{BiLSTM}}([\mathbf{C}, \mathbf{O}_{para}]) \in \mathbb{R}^{m \times d_2} \label{eq:sent} \\
	\mathbf{O}_{sent}&=\sigma(\mathbf{W}_s[\mathbf{M}_{s}[S^{(i)}_{s}]; \mathbf{M}_{s}[S^{(i)}_{e}]]) \label{eq:sent pred} 
\end{align}
where $S^{(i)}_{s}$ and $S^{(i)}_{e}$ denote the start and end token index of $i$-th sentence. The answer span and type can be predicted in a similar way.
\begin{align}
	\mathbf{O}_{start}&={\rm \textmd{SubLayer}}([\mathbf{C}, \mathbf{O}_{sup}]) \\
	\mathbf{O}_{end}&={\rm \textmd{SubLayer}}([\mathbf{C}, \mathbf{O}_{sup}, \mathbf{O}_{start}]) \\
	\mathbf{O}_{type}&={\rm \textmd{SubLayer}}([\mathbf{C}, \mathbf{O}_{sup}, \mathbf{O}_{end}])
\end{align}
where $\textmd{SubLayer}()$ denotes the similar LSTM layer and linear projection in eq. \ref{eq:sent}-\ref{eq:sent pred}. We compute a cross entropy loss over each outputs and jointly optimize them.
\begin{equation}
\begin{aligned}
	\mathcal{L}_{reader}=\lambda_{a}\mathcal{L}_{start} &+ \lambda_{a}\mathcal{L}_{end} + \lambda_{p}\mathcal{L}_{para} \\
	&+ \lambda_{s}\mathcal{L}_{sup} + \lambda_{t}\mathcal{L}_{type}
\end{aligned}
\end{equation}
We assign each loss term a coefficient.

\begin{table*}
	\small
	\centering
	\begin{tabular}{lcccc|cccc}	
		\toprule
		~ & \multicolumn{4}{c}{Full wiki} & \multicolumn{4}{c}{Distractor} \\
		\multirow{2}*{\textbf{Model}} & \multicolumn{2}{c}{\textbf{Answer}} & \multicolumn{2}{c}{\textbf{Sup}} & \multicolumn{2}{c}{\textbf{Answer}} & \multicolumn{2}{c}{\textbf{Sup}} \\
		\cmidrule{2-9}
		~ & \textbf{EM} & \textbf{F1} & \textbf{EM} & \textbf{F1} & \textbf{EM} & \textbf{F1} & \textbf{EM} & \textbf{F1} \\
		\midrule
		Baseline \cite{yang-etal-2018-hotpotqa} & 24.7 & 34.4 & 5.3 & 41.0 & 44.4 & 58.3 & 22.0 & 66.7 \\
		QFE \cite{nishida-etal-2019-answering} & - & - & - & - & 53.7 & 68.7 & 58.8 & 84.7 \\
		DFGN \cite{qiu-etal-2019-dynamically} & - & - & - & - & 55.4 & 69.2 \\
		Cognitive Graph QA \cite{ding-etal-2019-cognitive} & 37.6 & 49.4 & 23.1 & 58.5 & - & - & - & -  \\
		GoldEn Retriever \cite{qi-etal-2019-answering} & - & 49.8 & - & 64.6 & - & - & - & - \\
		SemanticRetrievalMRS \cite{nie-etal-2019-revealing} & 46.5 & 58.8 & 39.9 & 71.5 & - & - & - & - \\
		Transformer-XH \cite{Zhao2020Transformer-XH} & 50.2 & 62.4 & 42.2 & 71.6 & - & - & - & - \\
		GRR w. BERT \cite{Asai2020Learning} & 60.5 & 73.3 & 49.3 & 76.1 & 68.0 & 81.2 & 58.6 & 85.2\\
		\midrule
		DDRQA w. ALBERT-xxlarge$^{\clubsuit}$ \cite{zhang2020ddrqa}  & 62.5 & 75.9 & 51.0 & 78.8 & - & - & - & -\\
		\midrule
		Gated Memory Flow & \textbf{63.6} & \textbf{76.5} & \textbf{54.5} & \textbf{80.2} & \textbf{69.6} & \textbf{83.0} & \textbf{64.7} & \textbf{89.0}\\
		\bottomrule
	\end{tabular}
	\caption{\label{table:fullwiki}
	Results on HotpotQA development set in the fullwiki and distractor setting. ``-'' denotes no results are available. ``$\clubsuit$'' indicates the the work is recently presented on arXiv.}
\end{table*}

\section{Experiments}
In this section, we describe the setup and results of our experiments on the HotpotQA dataset. We compare our method with published state-of-the-art methods\cite{Asai2020Learning} in retrieval and downstream QA performance to demonstrate the superiority of our proposed architecture.
\subsection{Experimental Setup}
\paragraph{Dataset.}
Our experiments are conducted on HotpotQA \cite{yang-etal-2018-hotpotqa}, a widely used textual multi-hop question answering dataset. For each question, models are required to extract a span of text as an answer and corresponding sentences as supporting facts. 
Besides, there are two different settings in HotpotQA: distractor setting and full wiki setting. For each question, two evidence paragraphs and eight distractor paragraphs collected by TF-IDF are provided in the distractor setting. For the full wiki setting, each answer and supporting facts should be extracted from the entire Wikipedia. HotpotQA contains two question types: bridge question and comparison question. The former requires models to reasoning from one evidence to another. To answer the comparison question, models need to compare two entities described in two paragraphs.
\paragraph{Metrics.} We evaluate our pipeline system not only on downstream QA performance but also on the intermediate retrieval result. We report F1 and EM scores for evaluating QA performance and Supporting Fact F1 (Sup F1) and Supporting Fact EM (Sup EM) to evaluate the supporting fact sentences retrieval performance. In addition, joint EM and F1 scores are used to measure the system of the joint performance of the QA model.
To measure the intermediate retrieval performance, we follow Asai~\shortcite{Asai2020Learning} to use the three metrics: Answer Recall (AR), which evaluate whether the answer is located in the selected paragraphs, Paragraph Recall (PR), which evaluate if at least one of the evidence paragraphs is in retrieved set, Paragraph Exact Match (P EM), which evaluate whether all evidence paragraphs are retrieved. The number of selected paragraphs for the reader is alternative, hence we also report the precision score to evaluate the retrieval performance more appropriately.

\paragraph{Implementation Details.}
Considering accuracy and computational cost, we use different pre-trained models in different components in the system. We report the whole pipeline system result in Section \ref{sec:main} when the GMF is based on a RoBERTa-large model \cite{liu2019roberta}. In analysis experiments, the GMF is based on a RoBERTa-base model for saving computation cost. 
We use ALBERT-xxlarge \cite{lan2019albert} for the reader model. Furthermore, all hyper-parameters in our system are listed in Table \ref{table:hyper}.

\begin{table}
	\small
	\centering
	\begin{tabular}{lc|lc}
		\toprule
		\multicolumn{2}{c|}{\textbf{Retrieval model}} & \multicolumn{2}{|c}{\textbf{Reader Model}} \\
		\midrule
		$N_{kwm}$ & 10 & $\lambda_a, \lambda_p, \lambda_t$ & 1 \\
		$N_{tfidf}$ & 5 & $\lambda_s$ & 15 \\
		$h$ & 16 & epoch & 4 \\
		$d$ & 1024 & lr & 1e-5 \\
		$N_{retri}$ & 8 & batch & 8 \\
		$h_{d}$ & 0.025 & & \\
		gate value & 0.2 & & \\
		epoch & 2 & & \\
		lr & 1e-5 & & \\
		batch & 6 & & \\
		\bottomrule
	\end{tabular}
	\caption{\label{table:hyper}
	List of hyper-parameters used in our model.}
\end{table}

\begin{table}
	\small
	\centering
	\begin{tabular}{lccc}	
		\toprule
		\textbf{Methods} & \textbf{AR} & \textbf{PR} & \textbf{P EM} \\
		\midrule
		Entity-centric IR & 63.4 & 87.3 & 34.9 \\
		Cognitive Graph & 76.0 & 87.6 & 57.8 \\
		Semantic Retrieval & 77.9 & 93.2 & 63.9 \\
		GRR w. BERT & 87.0 & 93.3 & 72.7 \\
		\midrule
		GMF w. BERT & 90.8 & 94.1 & 85.7 \\
		GMF w. RoBERT & \textbf{91.7} & \textbf{94.7} & \textbf{86.3} \\
	\bottomrule
	\end{tabular}
	\caption{\label{table:retrieval}
	Comparing our retrieval method with other published methods across Answer Recall, Paragraph Recall and Paragraph EM metrics.}
\end{table}

\begin{table*}
	\small
	\centering
	\begin{tabular}{lcccc|cc}	
		\toprule
		\multirow{2}*{\textbf{Settings}} & \multicolumn{4}{c}{\textbf{Retrieval}} & \multicolumn{2}{c}{\textbf{QA}} \\
		~ & \textbf{AR} & \textbf{PR} & \textbf{P EM} & \textbf{Prec.} & \textbf{Joint EM} & \textbf{Joint F1} \\
		\midrule
		Gated Memory Flow & 91.7 & 94.7 & 86.3 & 49.1 & 39.6 & 65.8 \\
		\quad-Bi-direct. Doc. & 87.2 & 95.6 & 78.8 & 51.3 & 37.1 & 62.3 \\
		\quad-Threshold Gate & 91.2 & 94.3 & 85.6 & 33.8 & 39.0 & 65.3 \\
		\quad-Rewritable Memory & 91.4 & 94.5 & 85.8 & 42.5 & 39.1 & 65.1 \\
		\quad-Harder Negatives & 91.5 & 94.5 & 86.0 & 39.2 & 38.4 & 64.8 \\
	\bottomrule
	\end{tabular}
	\caption{\label{table:ablation}
	Ablation study on the effectiveness of the GMF model on the dev set in the full wiki setting.}
\end{table*}

\subsection{Main Results}
\label{sec:main}
\noindent\textbf{Downstream QA.}
We first evaluate our pipeline system on the downstream task on HotpotQA development set\footnote{We will also submit our model to blind test set soon.}. Table \ref{table:fullwiki} shows the results in full wiki setting. The proposed GMF outperforms all published works on every metric by a large margin. 
The GMF achieves 3.1/3.2 points absolute improvement on Answer EM/F1 scores and 5.2/4.1 on Sup EM/F1 scores over the published state-of-the-art results,  
which implies the superiority of the whole pipeline design. We also report our method results in the distractor setting. 
Our GMF also achieves strong results in this setting.

\noindent\textbf{Retrieval.}
The retrieval results in full wiki setting are presented in Table \ref{table:retrieval}. 
To fairly compared with the published state-of-the-art results, we also report the results of our model based on BERT.
Our GMF achieves 3.4 AR and 13.0 P EM over the previous state-of-the-art model. The significant improvement comes from the proposed architecture for solving multi-hop retrieval. The new framework abandons many unnecessary presuppositions in graph-based methods and does not retrieve a reasoning path explicitly. Our GMF benefits from the generalization and flexibility of the proposed framework, leading to a high recall of evidence paragraphs. Note that we achieve a high recall with very few retrieved paragraphs. The average number of retrieved paragraphs for each question is less than 4.

\subsection{Ablation Study}
\label{sec:ablation}
To evaluate the effectiveness of our proposed method, we perform ablation study on our GMF model. 
In table \ref{table:ablation}, we report the ablation results in the development set. From the table, we can see that only retrieved paragraphs connected from paragraphs in $P_{kwm}$ and $P_{tfidf}$ will improve the precision but significantly hurt the recall scores and downstream performance. 
We find different components do not influence the recall but lead to higher precision. 
In particular, using the external rewritable memory can provide 6.6\% precision improvement, which implies the effectiveness of modeling the history of retrieval. The gating mechanism can help the model not memorizes and retrieve irrelevant information. The precision will decreases by 16\% point after removing the gate. In addition, harder negatives also improve retrieval performance. The QA performance also consistently decreases after ablating each component.

\subsection{Analysis}
\label{sec:analysis}
\paragraph{Effectiveness of Negatives.} We find that the training data sampled from different sources will significantly affect the final retrieval performance. In Table \ref{table:training}, we report our GMF retrieval performance compared with a RoBERTa based re-rank baseline model, where the training data is sampled from $P_{kwm}$, $P_{tfidf}$, $P_{hyper}$ and paragraphs selected by a pre-trained model.

The experiments show that the models can achieve comparable recall in different settings. However, Comparing data sampled from $P_{kwm}$ and $P_{hyperlink}$, using training data sampled from $P_{tfidf}$, the GMF can gain about 2.4\% and 4.6\% precision scores improvement respectively.
The results imply that, in the training phase, examples sampled from $P_{tfidf}$ or $P_{kwm}$ are more confusing than $P_{hyperlink}$ for more lexical overlap. Therefore, we further leverage a neural model to rank the candidate paragraphs and sample the most challenging examples, leading to 9.9\% absolute precision improvement. In addition, our GMF achieves better retrieval performance in each different setting, 
which implies the Effectiveness of our proposed model.

\begin{table}
	\small
	\centering
	\begin{tabular}{llcccc}	
		\toprule
		\textbf{Model} & \textbf{Source} & \textbf{AR} & \textbf{PR} & \textbf{P EM} & \textbf{Prec.} \\
		\midrule
		baseline & K.W.M. & 92.2 & 94.6 & 86.4 & 37.3 \\
		baseline & TF-IDF & 91.8 & 94.5 & 86.3 & 39.2 \\
		baseline & Hyperlink & 91.3 & 94.6 & 86.4 & 32.2 \\
		baseline & BERT & \textbf{92.4} & \textbf{94.6} & \textbf{86.5} & \textbf{43.4} \\
		\midrule
		GMF & K.W.M. & 91.6 & 94.4 & 86.1 & 36.8 \\
		GMF & TF-IDF & 91.5 & 94.2 & 86.0 & 39.2 \\
		GMF & Hyperlink & 91.2 & 94.5 & 85.9 & 34.6 \\
		GMF & BERT & \textbf{91.7} & \textbf{94.7} & \textbf{86.3} & \textbf{49.1} \\
		\bottomrule
	\end{tabular}
	\caption{\label{table:training}
	Effectiveness of training data with different data distribution.}
\end{table}

\begin{figure*}
	\centering
	\includegraphics[width = 15cm]{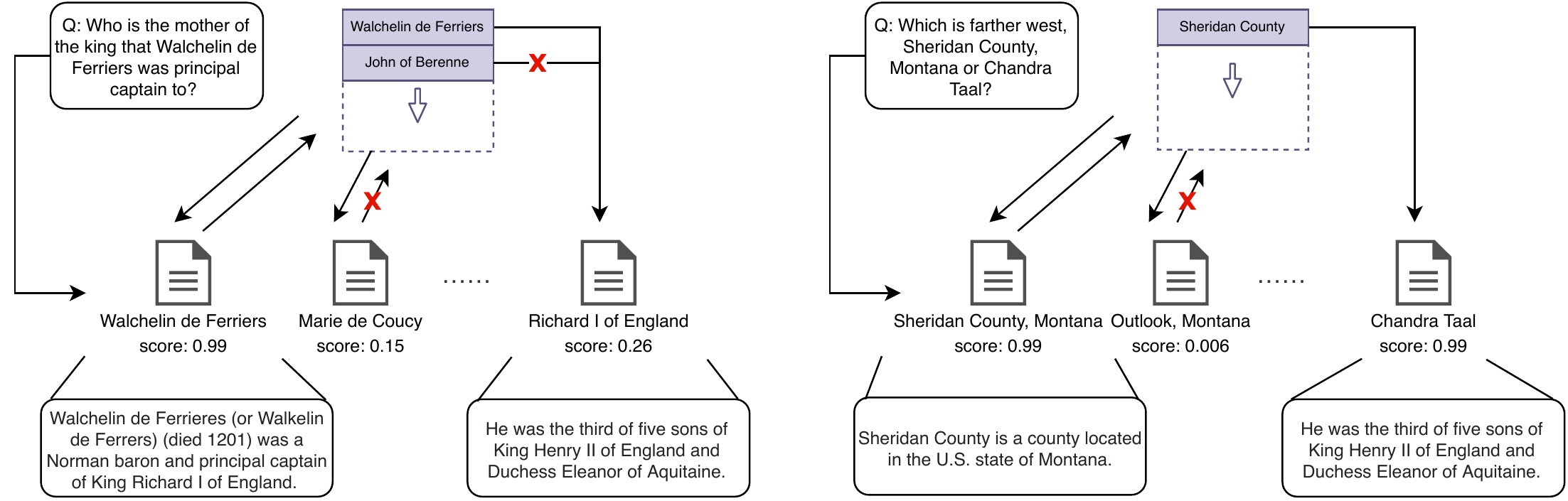}
	\caption{\label{fig:case}
	Examples of evidence paragraphs prediction in the full wiki setting of HotpotQA .
	}
\end{figure*}

\paragraph{Analysis on Candidate Set.} 
We also evaluate how different sizes of the candidate set to influence the downstream QA performance. Previous work 
\cite{Asai2020Learning} 
uses recall as a metric to measure the quality of retrieval. Intuitively, more retrieved paragraphs will cause a higher recall. However, does higher recall leads to better downstream task performance? We increase the number of paragraphs retrieved by the TF-IDF method in the term-based retrieval phase. 
As expected, Table \ref{table:different} shows more retrieved paragraphs lead to a higher recall. 
Notice that the number of candidate paragraphs $|P_{cand}|$ significantly increases with $N_{tfidf}$. 
Nevertheless, Joint EM and Joint F1 scores do not increase as recall. Because there are too many noise paragraphs in the candidate set, some will be retrieved and misleading the reader model. 
The analysis experiment suggests that only using recall as a metric to evaluate retrieval quality is not enough. We hope the future works introduce additional metrics (e.g., precision) to evaluate the retrieval model they proposed comprehensively.

\subsection{Case Study}
We provide two examples to understand how our model works. In Figure \ref{fig:case}, the first question is a bridge question. To answer the question, our model first retrieves the paragraph \textit{``Walchelin de Ferrieres''} with high confidence. Then the model takes paragraph \textit{``Marie de Councy''} and gives a 0.15 relevance score. The score is lower than the threshold value of the gate, so the paragraph will not be written to the memory. There are two paragraphs stored in the memory when the model is trying to calculate the relevance between the paragraph \textit{``Richard I England''} and the question. The model takes the representation of paragraph as input to address necessary information from the memory module, then the memorized paragraph \textit{``Walchelin de Ferriers''} is readout. 

The second case is a comparison question. The question asks to compare the locations of \textit{``Sheridan County, Montana''} and \textit{``Chandra Taal''}. 
The two evidence paragraphs gain a very high relevant score, while others score approximately equal to zero. 
In fact, because the keywords \textit{``Sheridan County, Montana''} and \textit{``Chandra Taal''} appear in the question, such comparison questions are comparatively easy for not only GMF but also a RoBERTa based re-rank baseline model. 
However, such questions may be tricky for graph-based retriever due to lack of hyperlink connection or the same entity mentions between them. In contrast, our architecture does not require such presuppositions, leading to better generalization and flexibility.

\begin{table}
	\small
	\centering
	\begin{tabular}{lccccc}	
		\toprule
		\textbf{Top $N_{tfidf}$} & \textbf{Recall} & \textbf{Num.} & \textbf{Joint EM} & \textbf{Joint F1} \\
		\midrule
		Top 5 & \textbf{93.7} & \textbf{94} & \textbf{40.1} & \textbf{66.1} \\
		Top 10 & 95.0 & 130 & 39.8 & 65.8 \\
		Top 15 & 95.7 & 167 & 39.5 & 65.6 \\
		Top 20 & 96.2 & 203 & 39.1 & 65.1 \\
	\bottomrule
	\end{tabular}
	\caption{\label{table:different}
	Comparing the performance with different recall of candidate set.}
\end{table}

\section{Conclusion}
\label{sec:conclusion}
In this paper, 
to tackle the multi-hop information retrieval challenge, 
we introduce an architecture that models a set of paragraphs as sequential data and iteratively identifies them. 
Specifically, 
we propose Gated Memory Flow to iterative read and memorize reasoning required information without noise 
information 
interference. We evaluate our method on both full wiki and distractor settings on the HotpotQA dataset and the method outperforms previous works by a large margin. In the future, we will attempt to design a more complicated model to improve retrieval performance and explore more about the effect of training data with different data distribution for multi-hop information retrieval. 

\bibliography{anthology}
\bibliographystyle{acl_natbib}

\end{document}